\documentclass[sigconf]{acmart}

\usepackage{amssymb} 
\usepackage{booktabs} 
\usepackage{multirow} 
\usepackage{colortbl} 
\usepackage{xcolor}   
\usepackage{algorithm} 
\usepackage{algpseudocode} 
\usepackage{tcolorbox}
\usepackage{tikz}
\usepackage{comment}
\usepackage{tablefootnote}
\usepackage{graphicx}
\usepackage{subcaption}
\usepackage{tabularx}
\usepackage{enumitem}
\usetikzlibrary{positioning, shapes.geometric, arrows.meta, shadows, calc, backgrounds, fit}

\definecolor{lightgrayline}{gray}{0.75}

\AtBeginDocument{%
  }

\copyrightyear{2026}
\acmYear{2026}
\setcopyright{cc}
\setcctype{by-nc-nd}
\acmConference[KDD '26]{Proceedings of the 32nd ACM SIGKDD Conference on Knowledge Discovery and Data Mining V.2}{August 09--13, 2026}{Jeju Island, Republic of Korea}
\acmBooktitle{Proceedings of the 32nd ACM SIGKDD Conference on Knowledge Discovery and Data Mining V.2 (KDD '26), August 09--13, 2026, Jeju Island, Republic of Korea}
\acmDOI{10.1145/3770855.3818912}
\acmISBN{979-8-4007-2259-2/2026/08}

\begin{document}

\title{MetaRanker: Human-in-the-loop Active Ranking for Metalens Image Quality}

\author{Yujin Park}
\orcid{0009-0001-8988-5698}
\affiliation{%
  \institution{Hanyang University}
  \city{Seoul}
  \country{Republic of Korea}
}
\email{yujin1019a@hanyang.ac.kr}

\author{Haejun Chung}
\authornote{Corresponding author.} 
\orcid{0000-0001-8959-237X}
\affiliation{%
  \institution{Hanyang University}
  \city{Seoul}
  \country{Republic of Korea}
}
\email{haejun@hanyang.ac.kr}

\author{Ikbeom Jang}
\authornotemark[1] 
\orcid{0000-0002-6901-983X}
\affiliation{%
  \institution{Hankuk University of Foreign Studies}
  \city{Yongin}
  \country{Republic of Korea}
}
\email{ijang@hufs.ac.kr}

\renewcommand{\shortauthors}{Yujin Park, Haejun Chung, and Ikbeom Jang}

\begin{abstract}

Image quality in modern imaging systems emerges from the coupled effects of the sensor, optics, and computational reconstruction. Ultra-thin metalenses offer a path toward substantial miniaturization, but practical designs often exhibit pronounced chromatic and field-dependent aberrations that decouple standard fidelity metrics (e.g., PSNR, SSIM) from human preference and downstream utility---a phenomenon we term \textit{recognizability collapse}. We introduce MetaRanker, a human-in-the-loop active ranking framework that formalizes metalens image quality in terms of semantic interpretability, defined as the degree to which humans can reliably recognize objects and structures in the presence of optical artifacts. 
MetaRanker combines a probabilistic preference model with uncertainty-aware query selection and leverages vision--language models (VLMs) to provide lightweight semantic priors. These priors are used only to guide the sampling of informative comparisons; human judgments remain the primary supervision signal throughout. Crucially, the system is self-correcting.
Across real-world and synthetic metalens datasets, MetaRanker reduces the annotation cost by approximately 80\% relative to exhaustive evaluation while achieving the best rank-aggregated inter-rater reliability. Our results confirm that standard metrics fail to capture interpretability in this domain, positioning MetaRanker as a practical step toward perceptually grounded metalens evaluation and future co-design.

\end{abstract}



\maketitle




\section{Introduction}
Modern imaging systems have evolved through advances in sensors and optics, yet a critical gap remains between physical fidelity and perceptual quality. While neural image signal processing (ISP) has bridged this gap for conventional cameras, the misalignment between distortion metrics (e.g., PSNR) and human utility is particularly acute for metalenses. Despite offering a scalable, planar form factor, metalenses suffer from severe chromatic and off-axis aberrations~\cite{kim2023scalable,yeo2025eigencwd}. Current reconstruction pipelines typically rely on proxy objectives that fail to capture \emph{semantic interpretability}—whether a human can reliably recognize objects amidst these unique artifacts. We observe a "recognizability collapse" where standard metrics correlate poorly with actual human visual task performance.~\cite{song2021ie,li2025image}

Bridging this gap requires treating human judgment as the ground-truth signal. However, collecting pairwise human preferences at scale is prohibitively expensive ($O(N^2)$), creating a significant data bottleneck for AI-driven scientific discovery. For example, even a modest set of $N=30$ images requires $N(N-1)/2=435$ exhaustive pairwise comparisons, whereas our protocol uses a fixed budget of only $B=90$ comparisons per session. To iterate effectively, metalens research needs a method to acquire high-quality preference signals under strict budget constraints.

To address this, we introduce \textbf{MetaRanker}, a budget-efficient active ranking framework. MetaRanker defines image quality via semantic interpretability and combines a probabilistic rating model with an uncertainty-aware acquisition strategy to prioritize informative decision boundaries. To mitigate cold-start inefficiency, we utilize a vision-language prior as a sampling guide—biasing early candidates without treating the prior as ground truth. Validated on the DRMI and Metaformer datasets, our approach achieves superior inter-rater reliability with a fraction of the annotation effort. Our contributions are:

\begin{itemize}[leftmargin=*]
    \item We introduce \textbf{MetaRanker}, a human-in-the-loop active ranking framework for metalens image quality.
    \item We propose an \textbf{uncertainty-aware acquisition function} to obtain human-preference labels only for decision-relevant pairs.
    \item We incorporate a \textbf{VLM-based semantic prior} to guide early sampling, effectively reducing cold-start waste without compromising ground-truth integrity.
    \item We demonstrate \textbf{budget-efficient performance}, achieving reliable preference supervision under limited budgets across diverse metalens datasets.
\end{itemize}

\begin{figure}[t]
    \centering
    \includegraphics[width=1.0\linewidth]{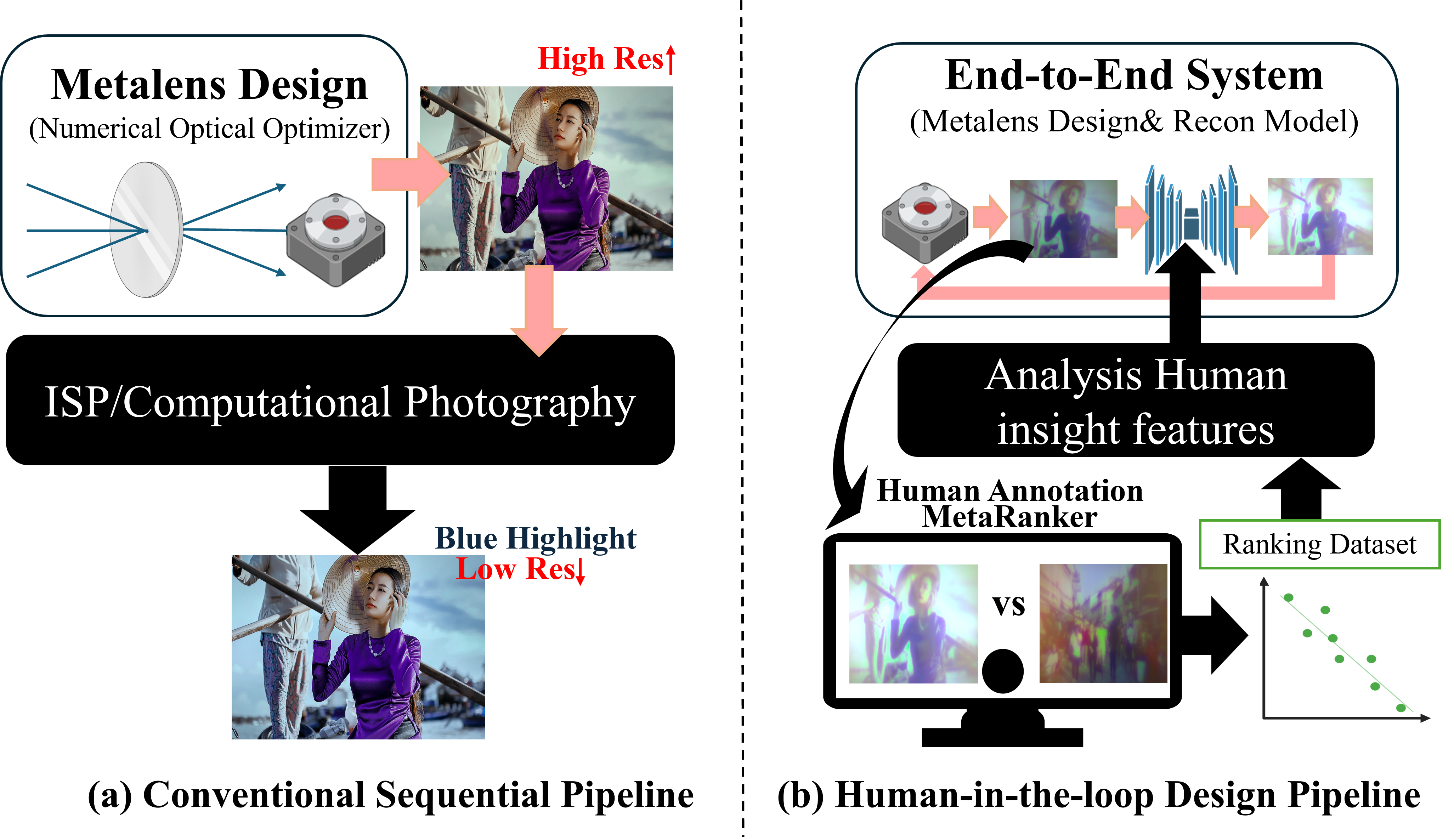}
    \caption{\textbf{Objective misalignment across the imaging pipeline.}
Proxy metrics can improve while human interpretability degrades; our ranking annotations enable a human-aligned signal that can be integrated into future closed-loop co-design.}
    \label{fig:pipeline_concept}
\end{figure}


\section{Related Work}
\label{sec:related}
\textbf{Metalens Imaging and the Perception--Distortion Trade-off.}
Metalenses implement a desired wavefront by patterning arrays of subwavelength resonators (meta-atoms), enabling an ultra-thin, planar form factor. However, because a metalens must jointly provide optical power and dispersion control, it is subject to more stringent bandwidth, aperture, and efficiency constraints than conventional multi-element refractive lens assemblies, which distribute aberration correction across multiple surfaces. In particular, enforcing the required phase profile over a broad, continuous spectral band (e.g., the visible) at practical apertures (e.g., millimeter to centimeter scale) generally demands large, spatially varying group delays that are difficult to realize within a thin metasurface and are bounded by fundamental delay--bandwidth and efficiency limits. 

As a result, many designs either accept a wavelength-dependent focal length, as in diffractive optics, or attempt to maintain an approximately wavelength-invariant focal length over a limited band via dispersion engineering, often at the cost of reduced bandwidth, numerical aperture, or focusing efficiency. The former inherently introduces strong \emph{chromatic aberration} through wavelength-dependent defocus. Moreover, due to field-angle sensitivity and limited degrees of freedom in a single surface, practical metalenses can exhibit substantial \emph{off-axis aberrations} (e.g., coma and astigmatism), producing spatially varying blur that worsens toward the image periphery. Consequently, practical metalens cameras often rely on computational reconstruction to recover semantically relevant details, making the end-to-end pipeline (optics $\rightarrow$ sensor $\rightarrow$ reconstruction) the appropriate unit of evaluation~\cite{khorasaninejad2017metalenses,hu2024metasurface}.

Recent approaches have advanced this restoration front, ranging from deep end-to-end learning~\cite{seo2024deep} to transformer-based aberration correction (e.g., MetaFormer)~\cite{lee2024aberration} and spatially varying deconvolution (e.g., EigenCWD)~\cite{yeo2025eigencwd}.
However, a key lesson from computational photography is that optimizing for distortion-based fidelity (e.g., PSNR) does not necessarily enhance perceptual quality, a phenomenon formalized as the perception--distortion trade-off~\cite{blau2018perception}.
Learned metrics like LPIPS~\cite{zhang2018unreasonable} and neural ISP studies~\cite{schwartz2018deepisp,ignatov2017dslr} further demonstrate that deep feature consistency aligns better with human preference than pixel-wise error. Our work bridges this gap in the metalens domain by addressing the \emph{data bottleneck} in human-grounded supervision: we introduce a budget-efficient ranking pipeline that targets semantic interpretability and can provide human-aligned supervision for future reconstruction or design optimization.

\begin{figure}[tbh!]
    \centering
    \includegraphics[width=1.0\linewidth]{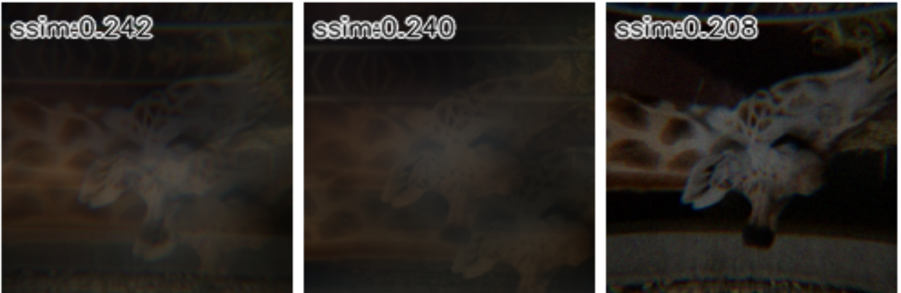}
        \caption{
        Metric--perception misalignment.
        Left images yield higher SSIM despite severe perceptual artifacts
        (e.g., chromatic aberration), illustrating that standard fidelity metrics can
        misalign with human-recognized image quality. Examples are adapted from
        DRMI~\cite{seo2024deep}, whose clean scenes are derived from DIV2K datasets
        ~\cite{agustsson2017ntire}.
        }
    \label{fig:metric_diff}
\end{figure}

\textbf{Subjective IQA and Pairwise Judgments.}
Subjective image quality assessment (IQA) is typically conducted via controlled protocols that collect human opinions under standardized viewing conditions.
While mean opinion score (MOS)~\cite{mikhailiuk2018psychometric} is common, pairwise comparisons are often preferred when absolute scoring is unreliable due to scale ambiguity or cognitive load.~\cite{xu2016pairwise}
Pairwise judgments can be converted into global rankings through probabilistic models such as Bradley--Terry~\cite{bradley1952rank} or Thurstone-style formulations, enabling robust aggregation under rater noise.

To further reduce the burden of data collection, recent works have explored progressive boosting~\cite{ling2020strategy,yang2021pair} and sorting-based protocols that lower comparison complexity to $\mathcal{O}(n\log n)$~\cite{maystre2017justsortit,jang2022decreasing}. While hybrid approaches like EZ-Sort~\cite{park2025ez} optimize this via zero-shot priors, such models are often brittle under domain shift, and logarithmic scaling remains prohibitive for large datasets. This necessitates budget-aware, uncertainty-driven active preference learning, particularly in noisy domains like metalens imaging; unlike fixed sorting schedules, this approach dynamically prioritizes only the most informative comparisons to maximize efficiency under a strict budget.

\textbf{Ranking Models with Uncertainty.}
Pairwise preference learning is closely related to classical paired-comparison models and modern skill-rating systems.
In online settings, rating systems such as Elo and the Glicko~\cite{glickman1995glicko, glickman2012example} family maintain per-item ratings with uncertainty, updating them sequentially with each comparison.
Uncertainty-aware models are attractive for human-in-the-loop pipelines because they can distinguish between ``confirmed low quality'' and ``insufficiently explored'' configurations.

\textbf{Task- and Semantics-Oriented Quality.}
Beyond perceptual pleasantness, an important line of work evaluates image quality through \emph{utility} for recognition, detection, or decision-making.
This perspective aligns with practical requirements in scientific imaging, where an image must be interpretable to support analysis. Accordingly, semantic interpretability provides a complementary axis to distortion- or perceptual-similarity metrics, particularly when the dominant failure mode is the loss of recognizable structures under severe degradations.

\textbf{Active Sampling for Preference Learning.}
Because pairwise labeling is expensive, active preference learning seeks to minimize comparisons by selecting informative pairs, often prioritizing high uncertainty or high expected information gain.~\cite{biyik2024batch}
GURO~\cite{bergstrom2024guro} provides a principled acquisition criterion that accounts for both epistemic and aleatoric uncertainty, enabling efficient preference learning with contextual attributes.
These ideas are directly relevant to budget-limited human ranking pipelines. MetaRanker builds on this direction while targeting metalens-specific artifacts and interpretability-driven objectives.

\begin{figure*}[t]
\centering
\includegraphics[width=1.0\textwidth]{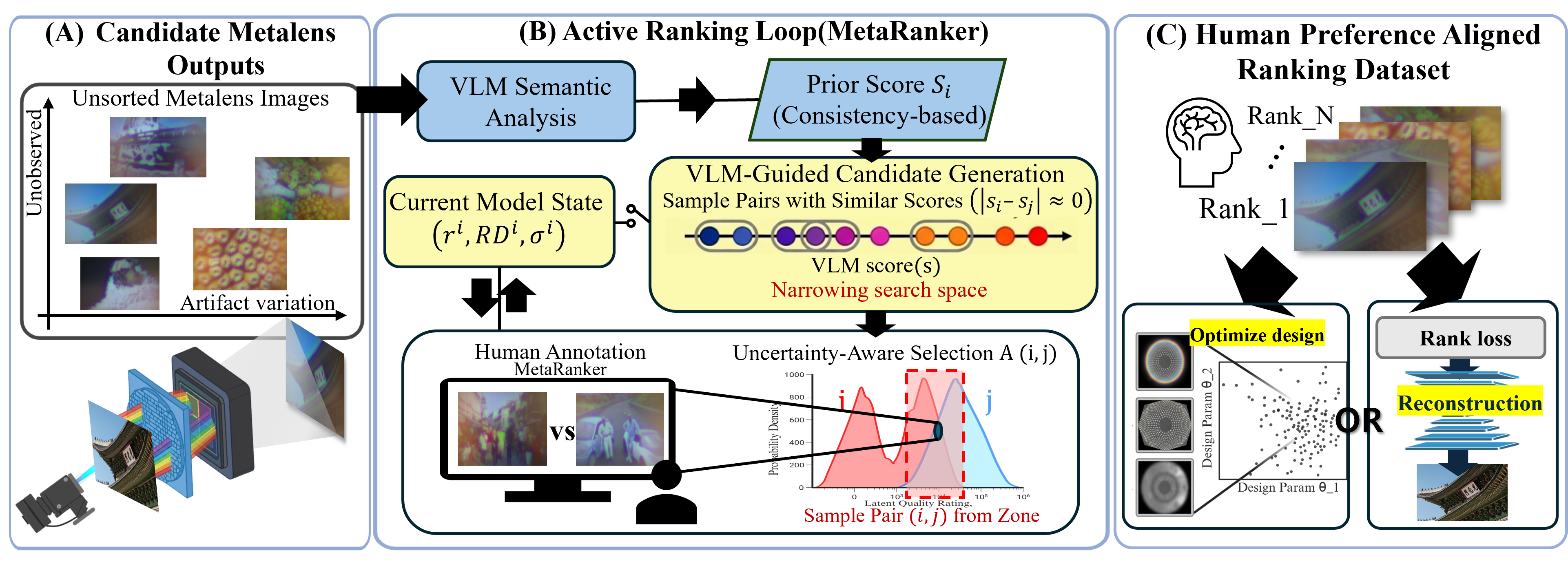}
\caption{\textbf{Overview of MetaRanker.}
(A) \textbf{Input:} Metalens outputs exhibit diverse, metric-misaligned artifacts (e.g., chromatic aberration, haze).
(B) \textbf{Active ranking:} A VLM-guided prior and uncertainty-aware sampling select informative pairs for human comparison.
(C) \textbf{Use:} The resulting human-aligned rankings provide evaluation ground truth and can support future reconstruction training (solid) or design optimization (dashed).}

\label{fig:main_overview}
\end{figure*}

\section{Methods}
\label{sec:methods}

\subsection{Problem Setup: Metalens Quality Ranking}
\label{sec:problem_setup}
Let $\mathcal{I}=\{1,\dots,N\}$ denote a set of metalens images generated under varying optical design parameters and reconstruction hyperparameters. Unlike natural images, metalens outputs suffer from unique degradations such as chromatic aberrations, radial blur, and reduced contrast, which often render traditional metrics (PSNR, SSIM) ineffective.
We define the ranking objective not merely as perceptual pleasantness, but as \textbf{semantic interpretability}: the ability of an observer to correctly identify objects and structures despite optical artifacts. We collect pairwise comparisons $(i,j,y)$ where $y \in \{0, 0.5, 1\}$ denotes the outcome: $y=1$ if $i$ is preferred, $y=0$ if $j$ is preferred, and $y=0.5$ indicates a tie. This explicitly handles ambiguity in recognizability under standardized annotation guidelines (Table~\ref{tab:rubric_checklist}). The goal is to infer a global utility score vector $\mathbf{r}\in\mathbb{R}^N$ while minimizing the annotation budget via active sampling.

\subsection{Probabilistic Ranking with Adaptive Volatility}
\label{sec:ranking_model} 

We adopt a probabilistic model based on Glicko-2~\cite{glickman2012example}, characterizing each image $i$ by a latent quality rating $r_i$ and a rating deviation (uncertainty) $\mathrm{RD}_i$. This formulation is particularly suitable for our setup because it explicitly quantifies epistemic uncertainty, allowing the system to distinguish between ``confirmed poor quality'' and ``unexplored parameters.''

\textbf{Pairwise Probability and Update.}
The probability that image $i$ is preferred over $j$ is modeled as:
\begin{equation}
\label{eq:glicko_prob}
\Pr(i\succ j) \;=\; \frac{1}{1 + 10^{-g(\mathrm{RD}_j)\,(r_i-r_j)/400}},
\end{equation}
where $q=\ln(10)/400$ and $g(\mathrm{RD}) = \left(1 + 3q^2\mathrm{RD}^2/\pi^2\right)^{-1/2}$. Upon observing outcome $y$, we update $r_i$ and $\mathrm{RD}_i$ following standard Glicko-2 procedures; ties are
encoded as $y{=}0.5$, yielding a near-zero update when the model already predicts near-equal quality, while a tie against an overconfident prediction still triggers the adaptive volatility mechanism below. Crucially, we incorporate an \textit{adaptive volatility} mechanism to handle the ``ambiguous artifacts'' often found in metalens imaging.

\textbf{Self-Correcting Property.}
Even when the VLM prior is imperfect and biases early sampling toward suboptimal regions, human feedback induces large prediction errors for inconsistent pairs.
Our adaptive volatility mechanism inflates the item's volatility $\sigma_i$ (and thus $\mathrm{RD}_i$) when the prediction error $e = |y - \Pr(i \succ j)|$ is high. This increases the chance that uncertain items are revisited by the acquisition function (Eq.~\ref{eq:acquisition}), enabling rapid recovery without manual intervention:
\begin{equation}
\label{eq:self_correct}
\sigma_i^{\text{new}} \;=\; \sigma_i + \alpha\cdot\max\bigl(0,\; |y - \Pr(i\succ j)| - \theta\bigr),
\end{equation}
where $\alpha$ controls the correction magnitude and $\theta$ is a dead-zone threshold that prevents unnecessary inflation from minor prediction errors.

\begin{algorithm}[t]
\caption{MetaRanker Active Ranking (default configuration)}
\label{alg:glicko2_active}
\begin{algorithmic}[1]
\Require Image set $\mathcal{I}$, VLM prior scores $s_i$, budget $B$
\State \textbf{Init:} $r_i \leftarrow 1500,\ \mathrm{RD}_i \leftarrow \mathrm{RD}_0,\ \sigma_i \leftarrow \sigma_0$ \Comment{uninformative init}

\For{$t=1$ to $B$}
    \State $\mathcal{C} \leftarrow \textsc{SampleCandidatePairs}(\mathcal{I}; s)$ \Comment{prior-guided candidate pool (hybrid)}
    \State $(i^*, j^*) \leftarrow \arg\max_{(i,j)\in\mathcal{C}}\ \mathcal{A}(i,j)$ \Comment{Eq.~\ref{eq:acquisition}}

    \If{\textsc{AutoOn} $\wedge$ $\textsc{Conf}(i^*,j^*) \ge \tau_{\text{auto}}$}
        \State $y \leftarrow \textsc{AutoLabel}(i^*,j^*;\ s)$ \Comment{weak supervision (optional)}
    \Else
        \State $y \leftarrow \textsc{QueryHuman}(i^*, j^*)$
    \EndIf

    \State $p \leftarrow \Pr(i^*\succ j^*)$ \Comment{before rating update}
    \State Update $(r_{i^*}, r_{j^*}, \mathrm{RD}_{i^*}, \mathrm{RD}_{j^*})$ by Glicko-2 using outcome $y$
    \State $\sigma_{i^*}, \sigma_{j^*} \leftarrow
    \sigma_{i^*}, \sigma_{j^*}
    + \alpha\cdot\max(0,\ |y-p|-\theta)$ \Comment{Eq.~\ref{eq:self_correct}}

\EndFor
\State \Return ranking by sorting $r_i$ (descending)
\end{algorithmic}
\end{algorithm}

\subsection{Uncertainty-Aware Active Sampling}
\label{sec:active_sampling}
To efficiently navigate the metalens design space under a fixed budget, we employ an acquisition function $\mathcal{A}(i, j)$ that combines uncertainty with a preference for near-boundary pairs (Fig.~\ref{fig:main_overview}(B)). Intuitively, MetaRanker targets comparisons with overlapping posteriors (a high-uncertainty region), which tend to be most informative for refining the ordering. For a candidate pair $(i, j)$, the score is defined as:

\begin{equation}
\label{eq:acquisition}
\mathcal{A}(i, j) = 
\underbrace{(\mathrm{RD}_i + \mathrm{RD}_j)}_{\text{Uncertainty}}
\cdot
\underbrace{\Bigl(1 + \frac{|r_i - r_j|}{\kappa}\Bigr)^{-1.5}}_{\text{Fine-grained Closeness}}
\cdot
\underbrace{\Bigl(1 + \lambda(c_i + c_j)\Bigr)^{-1}}_{\text{Novelty}},
\end{equation}
where $c_i$ is the comparison count, $\kappa=100$ is the closeness scaling factor, and $\lambda$ is the novelty decay weight.

\begin{table}[t]
\centering
\caption{Metalens image quality assessment guidelines for the pairwise annotation. Designed by a domain expert.}
\label{tab:rubric_checklist}
\footnotesize
\setlength{\tabcolsep}{4pt}
\renewcommand{\arraystretch}{1.15}
\begin{tabularx}{\columnwidth}{@{}>{\centering\arraybackslash}m{1.0cm}X@{}}
\toprule
\textbf{Step} & \textbf{Checklist} \\
\midrule
0 & \textbf{Category fixed:} Each session is conducted within a single category (Animal or Landscape); all pairs shown to annotators are pre-filtered accordingly. \\
1 & \textbf{Recognizable?} Which image lets you identify the main object/scene \emph{more confidently} (less ambiguity)? \\
2 & \textbf{(Animal) Texture cues:} Which shows clearer discriminative textures (fur/skin pattern, facial/limb contours) rather than just “looking sharp”? \\
& \textbf{(Landscape) Structure cues:} Which preserves coherent scene layout (boundaries/horizon/building edges) without breaking understanding? \\
3 & \textbf{Artifact penalty:} If both are recognizable, pick the one with fewer metalens artifacts that harm understanding (color fringing/field blur/haze). \\
4 & \textbf{Tie rule:} If still tied, prefer lower artifact severity; if both are unrecognizable, choose the “less collapsed” one (any stable structure). \\
\bottomrule
\end{tabularx}
\end{table}

\textbf{Optimization Rationale.}
Standard active learning often prioritizes pure uncertainty. However, in metalens optimization, many configurations yield clearly unusable (collapsed) images. Comparing two ``noise'' images is uninformative.
Our strategy introduces a sharpened closeness term with an exponent of $1.5$ and a scaling factor $\kappa=100$. This focuses the budget on the \textit{Pareto frontier} of the design space distinguishing between ``slightly blurry'' and ``sharp but aberrant'' images rather than wasting queries on obvious decisions or varying grades of noise.

\textbf{Theoretical Convergence.}
Active ranking is most informative when $\Pr(i \succ j) \approx 0.5$, where Fisher information is maximized.
With uninformative initialization, purely random sampling wastes budget on trivially separable pairs.
Our prior-guided \emph{candidate generation} increases the likelihood of sampling near-boundary comparisons early, while the Glicko-2 uncertainty terms drive subsequent refinement.
This converts the budget from global discovery to local boundary refinement, accelerating convergence under fixed annotation budgets.

\subsection{VLM-based Semantic Prior}
\label{sec:prior_init}
Cold-start ranking is inefficient because random initial pairs often have obvious quality differences. To accelerate convergence, we inject domain knowledge using an \textbf{off-the-shelf Vision-Language Model (VLM)}. While our framework is compatible with various foundational models, we employ LLaVA-7b~\cite{liu2023visual} in our implementation to estimate a proxy quality score $s_i$ before human annotation begins.
\begin{equation}
\label{eq:prior_score}
s_i = w_1 \cdot \bar{v}_i + w_2 \cdot \Omega_i + w_3 \cdot n_i,
\qquad
n_i =
\frac{\log(1 + \bar{N}_{\text{obj}})}
{\log(1 + N_{\max})},
\end{equation}
We use $N_{\max}=8$, map visibility labels as \texttt{clear}$\mapsto1.0$, \texttt{somewhat}$\mapsto0.5$, and \texttt{unclear}$\mapsto0.0$, and set $(w_1,w_2,w_3)=(1/3,1/3,1/3)$ unless otherwise stated, so $s_i\in[0,1]$.

\begin{figure}[t]
    \centering
    \begin{tcolorbox}[
        colback=gray!10!white,
        colframe=gray!50!black,
        title=\textbf{VLM Prompt for Quality Assessment},
        boxsep=2pt, left=4pt, right=4pt, top=3pt, bottom=3pt, arc=2pt
    ]
    \small 
    \textbf{System Instruction:} You analyze image recognizability and description stability. Return ONLY a JSON object.

    \vspace{4pt} 
    \textbf{Response Format:}
    {\ttfamily\small
    \{ \\
    \hspace*{1em} "caption": "1-2 sentences describing content", \\
    \hspace*{1em} "objects": ["list of clearly identifiable objects"], \\
    \hspace*{1em} "visibility": "clear | somewhat | unclear", "confidence": 0.0 - 1.0 \\
    \}
    }
    \vspace{4pt} 
    {\raggedright
    \textbf{Constraint:} If the image is blurry or unclear, set {visibility="unclear"} and keep the object list minimal.
    \par} 
    \end{tcolorbox}

    \vspace{-0.2cm} 
    \caption{The structured prompt used to extract semantic metadata from LLaVA. By enforcing a JSON output with an explicit object list, we enable quantitative consistency checks.}
    \label{fig:vlm_prompt}
\end{figure}

\textbf{Semantic Consistency as Quality Proxy.}
We hypothesize that high-quality metalens images should yield consistent semantic descriptions, whereas degraded images lead to hallucinations or failure to detect objects. For each image $i$, we prompt the VLM $K$ times with a fixed temperature of 0.7 to induce stochasticity, generating: (1) a list of detected objects $\mathcal{O}_k$, (2) a visibility label $v_k \in \{\text{clear, somewhat, unclear}\}$, and (3) a confidence level $c_k$.

We define the \textit{semantic consistency} $\Omega_i$ using the Jaccard index of detected objects across sampled runs:
\begin{equation}
\Omega_i = \frac{1}{K(K-1)} \sum_{k \neq m} \frac{|\mathcal{O}_k \cap \mathcal{O}_m|}{|\mathcal{O}_k \cup \mathcal{O}_m|}.
\end{equation}
A high $\Omega_i$ indicates that the optical features are distinct enough to be robustly recognized.

\textbf{Prior Score Formulation and Usage.}
We convert the VLM outputs into a scalar prior score $s_i \in [0,1]$ (Eq.~\ref{eq:prior_score}).
However, we empirically found that initializing rating values with $s_i$ introduces anchoring bias (see Section~\ref{sec:results_ablation}).
Therefore, we initialize all items with an uninformative rating $r_i^{(0)}=1500$ and a common uncertainty $\mathrm{RD}_i^{(0)}=\mathrm{RD}_0$.
The prior is used \emph{solely} to guide early \textbf{candidate pair generation}: we preferentially sample pairs with similar prior scores (or adjacent ranks under $s_i$), thereby increasing the probability of querying informative, near-boundary comparisons without committing to the prior as ground truth. This strategy efficiently mitigates the risk of hallucinated priors in specialized scientific domains, where off-the-shelf VLMs usually lack the domain-specific calibration required for reliable absolute scoring.

\textbf{Robustness to Imperfect VLM and Domain Gaps.}
The VLM prior may be imperfect because off-the-shelf VLMs are not calibrated for metalens artifacts. MetaRanker limits this risk by using the prior only for candidate generation, not as a label or rating initializer. When human feedback contradicts the current model prediction, adaptive volatility increases uncertainty and redirects later queries toward unresolved items (Eq.~\ref{eq:self_correct}).

\noindent\textbf{Optional Auto-comparison.}
To reduce human queries, we optionally auto-label a comparison when the model is highly confident (e.g., based on a large predicted margin or high prior confidence).
Auto-labeled outcomes are treated as weak supervision and can be disabled without changing the core ranking model.


\begin{figure}[t]
    \centering
    \includegraphics[width=1\linewidth]{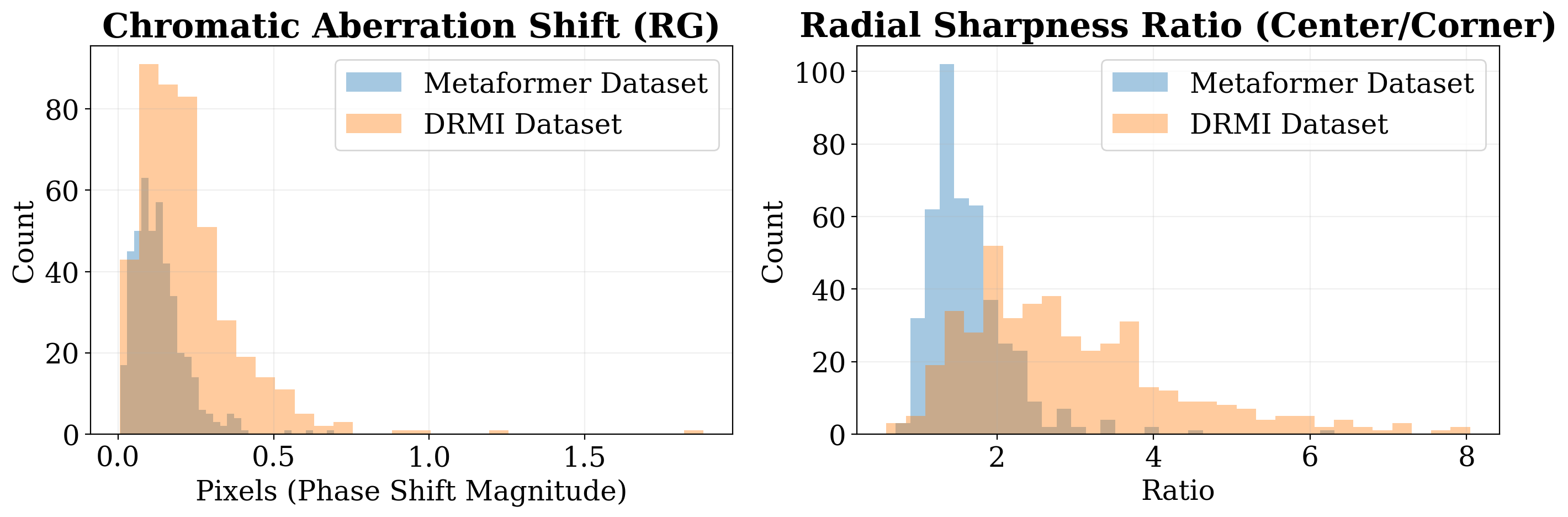}
    \caption{Dataset-level optical severity. CA Shift (px) measures chromatic misregistration, and radial sharpness ratio measures center-to-corner sharpness degradation.}
    \label{fig:severity_dataset}
\end{figure}

\section{Experiments}
\label{sec:experiments}

\subsection{Datasets}
\label{sec:datasets}
We evaluate MetaRanker on two metalens datasets with distinct degradation processes, and a controlled simulation used only for the human--metric agreement analysis.

\textbf{DRMI (real images).}
DRMI~\cite{seo2024deep} contains images directly captured through a custom fabricated metalens, exhibiting chromatic and field-dependent aberrations as well as sensor noise and real-world imperfections. Following the DRMI data acquisition protocol, the underlying clean scenes are cropped from DIV2K datasets~\cite{agustsson2017ntire}, displayed on a monitor, and captured through the metalens imaging system. We use DRMI as a real-capture benchmark and cite both the benchmark and the underlying source-image dataset.

\textbf{Metaformer (PSF-based synthesized images).}
Following the released MetaFormer pipeline~\cite{lee2024aberration}, degraded inputs are generated by convolving clean Open Images V7 source images~\cite{benenson2022colouring} with calibrated spatially varying PSFs and associated noise models, yielding more regular and controllable degradations than real capture.
We use this synthetic degradation setting because the private real-capture subset reported in the original work is not publicly accessible.

\textbf{Sim (PSF-Angle, controlled).}
To isolate optical degradation from content variability, we fix a single scene and convolve it with PSFs sampled at $N_{\text{sim}}{=}30$ uniformly spaced field angles (from on-axis to maximum field height), generating images with strictly monotonically increasing aberration severity. This controlled set is used exclusively for Table~\ref{tab:metric_vs_human_combined}. One annotator repeats multiple ranking runs on Sim while varying only the algorithmic seed; we report mean$\pm$std across runs.


\subsection{Setup and Evaluation}
\label{sec:protocol_stat}

For each dataset--category--method condition, we use the same $N{=}30$ item set and allocate a fixed budget of $B{=}90$ pairwise comparisons ($3N$) per annotator session. The evaluation used a custom web interface with randomized left/right positions, neutral background, and fixed display dimensions.
Human annotation was performed by seven annotators (one co-author and six independent experts in visual analysis and/or metalens imaging). Each annotator completed one independent $B=90$ session per dataset--category--method condition; all $B=90$ comparisons in the main benchmark were human-labeled, and auto-comparison was disabled. We compare MetaRanker against GURO~\cite{bergstrom2024guro}, Bayesian Comparative Judgment (Bayesian CJ)~\cite{gray2024bayesian}, and LBPS-EIC~\cite{mohammadi2025uncertainty} under the same budget constraint. We report inter-rater reliability (IRR) using Kendall's $\tau$ and Spearman's $\rho$; descriptive uncertainty summaries are provided in Appendix~\ref{app:full_stats}.
\noindent\textbf{Baselines.}
GURO~\cite{bergstrom2024guro} is a contextual preference learner that models pairwise outcomes with a feature-dependent logistic model and selects pairs by an information criterion in the learned embedding space.
Its performance therefore depends on whether the chosen visual features form a meaningful geometry for metalens-specific artifacts.
Bayesian CJ~\cite{gray2024bayesian} is a Bayesian Bradley--Terry-style model that maintains a per-item posterior (mean and uncertainty) and performs \emph{active} pair selection by prioritizing comparisons that are both uncertain (high posterior variance) and near the decision boundary (high $p(1-p)$), with additional novelty/balance terms.
We treat Bayesian CJ as a strong uncertainty-aware active-ranking baseline rather than a passive rank aggregation method.
LBPS-EIC~\cite{mohammadi2025uncertainty} is a learning-based pairwise comparison method that leverages perceptual embeddings and information-theoretic criteria for active pair selection; we include it as an additional competitive baseline following reviewer recommendation. To manage annotation cost in this specialized domain, we applied a two-stage screening procedure: all candidate methods were first evaluated via simulation, and the top-performing methods were advanced to the human annotation phase. 
For descriptive comparison between methods, we report paired non-parametric bootstrap confidence intervals and Cliff's $\delta$ effect sizes for Kendall's $\tau$ and Spearman's $\rho$ differences (Appendix~\ref{app:full_stats}).
To standardize judgments toward semantic interpretability (recognizability), annotators followed a concise guideline formulated by a domain expert (15 years of experience) grounded in established perceptual research~\cite{li2025image,redi2013crowdsourcing} (Table~\ref{tab:rubric_checklist}).

\begin{table*}[bht]
\centering
\begin{minipage}{\textwidth}
    \centering
    \caption{\textbf{Inter-rater reliability (IRR) of metalens image quality assessment.}
    The mean$\pm$std of pairwise session agreement is shown.
    Rank Sum is computed as Rank(Kendall)+Rank(Spearman) within each
    dataset--category condition; lower is better. Final Rank is determined
    by Rank Sum, with ties resolved by Kendall rank and then Spearman rank.
    Bold indicates the best metric value or best rank; underline indicates
    the second-best metric value.}
    \label{tab:human_benchmark_stacked}

    \small
    \setlength{\tabcolsep}{4pt}
    \renewcommand{\arraystretch}{1.15}

    \begin{tabular}{ll l c c c c}
        \toprule
        \textbf{Dataset} & \textbf{Category} & \textbf{Method}
            & \textbf{Kendall}
            & \textbf{Spearman}
            & \textbf{Rank Sum}$\downarrow$
            & \textbf{Final Rank}$\downarrow$ \\
        \midrule

        \multirow{8}{*}{\textbf{DRMI}~\cite{seo2024deep}}
        & \multirow{4}{*}{Animal}
            & GURO~\cite{bergstrom2024guro}
            & $0.426{\pm0.394}$ & $0.358{\pm0.596}$ & 8 & 4 \\
        &   & Bayesian CJ~\cite{gray2024bayesian}
            & $0.618{\pm0.240}$ & $\mathbf{0.846{\pm0.065}}$ & 4 & 2 \\
        &   & LBPS-EIC~\cite{mohammadi2025uncertainty}
            & $\underline{0.637{\pm0.228}}$ & $0.814{\pm0.072}$ & 5 & 3 \\
        &   & \textbf{Ours}
            & $\mathbf{0.695{\pm0.190}}$ & $\underline{0.830{\pm0.056}}$ & \textbf{3} & \textbf{1} \\
        \cmidrule(lr){2-7}

        & \multirow{4}{*}{Landscape}
            & GURO~\cite{bergstrom2024guro}
            & $0.423{\pm0.385}$ & $0.443{\pm0.752}$ & 8 & 4 \\
        &   & Bayesian CJ~\cite{gray2024bayesian}
            & $0.450{\pm0.337}$ & $0.633{\pm0.109}$ & 6 & 3 \\
        &   & LBPS-EIC~\cite{mohammadi2025uncertainty}
            & $\underline{0.773{\pm0.133}}$ & $\mathbf{0.912{\pm0.032}}$ & \textbf{3} & 2 \\
        &   & \textbf{Ours}
            & $\mathbf{0.786{\pm0.128}}$ & $\underline{0.859{\pm0.076}}$ & \textbf{3} & \textbf{1} \\
        \midrule

        \multirow{8}{*}{\textbf{MetaFormer}~\cite{lee2024aberration}}
        & \multirow{4}{*}{Animal}
            & GURO~\cite{bergstrom2024guro}
            & $0.272{\pm0.491}$ & $0.098{\pm0.893}$ & 8 & 4 \\
        &   & Bayesian CJ~\cite{gray2024bayesian}
            & $\underline{0.612{\pm0.236}}$ & $\underline{0.862{\pm0.043}}$ & 4 & 2 \\
        &   & LBPS-EIC~\cite{mohammadi2025uncertainty}
            & $0.508{\pm0.298}$ & $0.713{\pm0.103}$ & 6 & 3 \\
        &   & \textbf{Ours}
            & $\mathbf{0.794{\pm0.152}}$ & $\mathbf{0.897{\pm0.058}}$ & \textbf{2} & \textbf{1} \\
        \cmidrule(lr){2-7}

        & \multirow{4}{*}{Landscape}
            & GURO~\cite{bergstrom2024guro}
            & $\underline{0.783{\pm0.158}}$ & $\mathbf{0.935{\pm0.029}}$ & \textbf{3} & 2 \\
        &   & Bayesian CJ~\cite{gray2024bayesian}
            & $0.644{\pm0.334}$ & $0.854{\pm0.039}$ & 7 & 4 \\
        &   & LBPS-EIC~\cite{mohammadi2025uncertainty}
            & $0.657{\pm0.194}$ & $0.813{\pm0.072}$ & 7 & 3 \\
        &   & \textbf{Ours}
            & $\mathbf{0.840{\pm0.095}}$ & $\underline{0.877{\pm0.066}}$ & \textbf{3} & \textbf{1} \\
        \bottomrule
    \end{tabular}

\end{minipage}
\end{table*}

\begin{table}[t]
    \centering
    \caption{Human--metric agreement showing the failure of standard metrics.
    DRMI (diverse): Real-world images with varied scenes.
    Sim (controlled): Simulation with fixed content to isolate optical degradation.
    Agreement is consistently low across all metric types, including recent learned metrics
    (CLIP-IQA+, FGResQ), confirming that the metric--human gap cannot be closed by
    stronger automated IQA alone.}
    \label{tab:metric_vs_human_combined}
    
    \resizebox{\columnwidth}{!}{%
    \setlength{\tabcolsep}{4pt}
    \renewcommand{\arraystretch}{1.15}
    
    \begin{tabular}{llcccc}
        \toprule
        & & \multicolumn{2}{c}{\textbf{Diverse (DRMI)}} & \multicolumn{2}{c}{\textbf{Controlled (Sim)}} \\
        \cmidrule(lr){3-4} \cmidrule(lr){5-6}
        \textbf{Type} & \textbf{Metric} & \textbf{Kendall} & \textbf{Spearman} & \textbf{Kendall} & \textbf{Spearman} \\
        \midrule
        \multirow{5}{*}{\textbf{No-Reference}}
         & NIQE~\cite{mittal2012making}
             & $0.13{\pm}0.08$  & $0.18{\pm}0.12$
             & $0.28{\pm}0.07$  & $0.40{\pm}0.11$ \\
         & MUSIQ~\cite{ke2021musiq}
             & $0.05{\pm}0.03$  & $0.07{\pm}0.05$
             & $0.06{\pm}0.06$  & $0.09{\pm}0.08$ \\
         & MANIQA~\cite{yang2022maniqa}
             & $-0.06{\pm}0.07$ & $-0.09{\pm}0.10$
             & $0.06{\pm}0.07$  & $0.08{\pm}0.10$ \\
        & CLIP-IQA+~\cite{wang2023clipiqa}
             & $-0.05{\pm}0.06$ & $-0.07{\pm}0.09$
             & $-0.06{\pm}0.05$ & $-0.08{\pm}0.07$ \\
         & FGResQ~\cite{sheng2026fine}
             & $-0.32{\pm}0.11$ & $-0.45{\pm}0.14$
             & $-0.11{\pm}0.08$ & $-0.15{\pm}0.10$ \\
        \midrule
        \multirow{3}{*}{\textbf{Full-Reference}}
         & LPIPS~\cite{zhang2018unreasonable}
             & $-0.08{\pm}0.10$ & $-0.11{\pm}0.12$
             & $0.14{\pm}0.08$  & $0.19{\pm}0.10$ \\
         & SSIM~\cite{wang2004image}
             & $-0.08{\pm}0.09$ & $-0.11{\pm}0.11$
             & $0.09{\pm}0.09$  & $0.12{\pm}0.13$ \\
         & PSNR~\cite{huynh2008scope}
             & $-0.08{\pm}0.08$ & $-0.12{\pm}0.11$
             & $0.06{\pm}0.09$  & $0.08{\pm}0.12$ \\
        
        \bottomrule
    \end{tabular}%
    }
\end{table}

\section{Results}
\label{sec:results}
\subsection{Ranking Performance and Metric Alignment}
\label{sec:metric_ali}

\textbf{Inter-Rater Reliability under Fixed Budget}
\label{sec:results_irr}
Table~\ref{tab:human_benchmark_stacked} reports IRR and condition-wise rank aggregation under the fixed budget constraint ($B=3N$). MetaRanker achieves the highest Kendall's $\tau$ in all four conditions, ranging from $0.695$ (DRMI Animal) to $0.840$ (MetaFormer Landscape). When Kendall and Spearman ranks are aggregated, MetaRanker also obtains the best Final Rank in every dataset--category condition. For Kendall's $\tau$, descriptive bootstrap summaries (Appendix Table~\ref{tab:full_bootstrap}) show directionally positive
improvements in all $12$ baseline-condition comparisons (3 baselines $\times$ 4 conditions). Against Bayesian CJ, Kendall's $\tau$ gains are $+0.077$ on DRMI Animal, $+0.336$ on DRMI Landscape, $+0.182$ on Metaformer Animal, and $+0.196$ on Metaformer Landscape. Against GURO, all conditions show positive Kendall gains (e.g., DRMI Animal $\Delta\tau=+0.269$; DRMI Landscape $\Delta\tau=+0.363$).
In terms of annotation efficiency, the human benchmark uses $B=90$ comparisons instead of $435$ exhaustive pairs for $N=30$, a 79.3\% reduction, while preserving the observed Kendall and Final-Rank gains.

\textbf{IQA Metrics versus Human Rankings}
\label{sec:results_metrics}
Table~\ref{tab:metric_vs_human_combined} compares standard IQA metrics with reference-free human rankings. In DRMI (diverse scenes), full-reference metrics are near-zero or negative (e.g., PSNR/SSIM/LPIPS Kendall $\tau\approx -0.08$), indicating structural misalignment. In the controlled Sim (PSF-Angle) setting where content is fixed, alignment improves only modestly: NIQE reaches Kendall $\tau=0.28\pm0.07$ (Spearman $\rho=0.40\pm0.11$), while full-reference metrics remain low ($\tau<0.15$). This shows that the mismatch is not explained solely by content diversity. Figure~\ref{fig:metric_diff} illustrates representative cases where pixel-level similarity fails to reflect semantic interpretability under chromatic/field-dependent aberrations.

\subsection{Annotation Difficulty Structure across Conditions}
\label{sec:results_difficulty}
\begin{figure}
    \centering
    \includegraphics[width=1\linewidth]{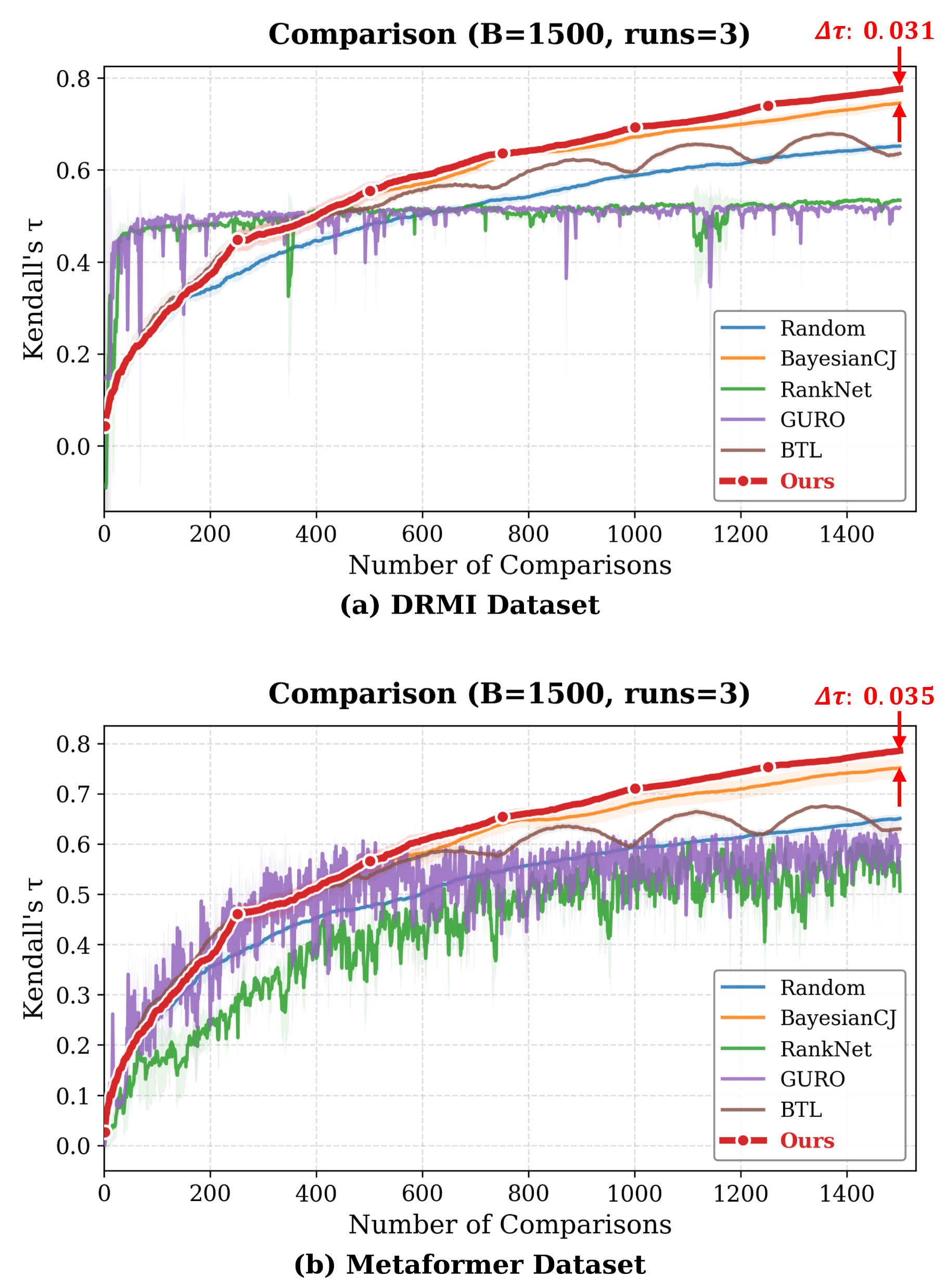}
    \caption{Controlled simulation against exhaustive PSNR ground truth ($N$=$500$, $B$=$1500$, 3 runs). Under this deterministic proxy target, Ours and Bayesian CJ converge similarly, indicating comparable search efficiency when the target ranking is internally consistent. This experiment evaluates algorithmic convergence only and does not validate PSNR as a perceptual quality target. In contrast, feature-based methods (GURO, RankNet) plateau at $\tau \leq 0.6$ on Metaformer, revealing the limitations of deep feature representations for metalens artifacts.}
        \label{fig:simulation_annotation}
\end{figure}

Figure~\ref{fig:severity_dataset} shows that DRMI exhibits markedly higher optical severity (CA shift and radial deviation) than Metaformer. Table~\ref{tab:difficulty_advantage} characterizes the annotation difficulty at the condition level.
DRMI Landscape exhibits a \emph{polarized} distribution (Easy 63.5\%, Hard 15.7\%), DRMI Animal is \emph{diffuse} (Easy 49.4\%, Hard 33.6\%), and both Metaformer conditions show \emph{plateau} distributions (Easy $>72\%$, Hard $<7\%$).
Notably, MetaRanker's gain over Bayesian CJ does not scale monotonically with overall optical severity or mean difficulty; the largest gain($+0.336$) occurs in the polarized regime.
Figure~\ref{fig:simulation_annotation} reports convergence against an exhaustive ground-truth ranking determined by PSNR on a larger set ($N{=}500$, $B{=}1{,}500$; 3 runs). 

We employ PSNR here solely as a deterministic oracle to benchmark search efficiency, isolating algorithmic convergence from the human label noise and perceptual misalignment discussed earlier.

Given this complete ground truth, MetaRanker and Bayesian CJ converge to similar levels ($\tau \approx 0.77$--$0.74$, $\Delta\tau < 0.03$), whereas feature-based methods (GURO, RankNet) converge substantially lower ($\tau \leq 0.6$ on Metaformer) and exhibit high run-to-run variance.
This performance gap is consistent with a separate feature space analysis (available in the released code repository): NSS features separate metalens outputs from the natural image manifold, but deep ResNet50 embeddings entangle metalens artifacts with standard distortions, providing no discriminative boundaries for quality-based active selection.


\subsection{Ablation Study}
\label{sec:results_ablation}

Table~\ref{tab:ablation_summary} in Appendix~\ref{app:ablation} reports ablation results under a controlled synthetic setup ($N{=}600$, $B{=}1{,}800$). H1 shows that the VLM prior is beneficial when used as a sampling guide rather than as a rating warm-start: the flat-start sampling-guide variant reaches $\tau=0.804$, improving over the flat-start baseline ($\tau=0.797$) by $\Delta\tau=+0.007$, whereas direct VLM-biased warm-starting decreases performance to $\tau=0.785$. H2 confirms the value of active acquisition, with the hybrid strategy improving over random selection by $\Delta\tau=+0.069$. H3 further shows that explicit rating uncertainty is essential: Glicko-2 with adaptive volatility outperforms Elo with $K=32$ by $\Delta\tau=+0.427$. Because these ablations use a synthetic Bradley--Terry oracle, we interpret them as mechanism checks supporting flat initialization, prior-guided candidate generation, active acquisition, and Glicko-2-style uncertainty tracking.

\begin{table}[t]
\centering
\caption{\textbf{Annotation difficulty structure.}
Easy/Middle/Hard: fraction of pairs by annotator agreement ($\geq$0.8 / between / $\leq$0.6).
Regime: qualitative summary of the difficulty distribution pattern.
The difficulty structure helps explain the varying effectiveness of active ranking across conditions.}
\label{tab:difficulty_advantage}
\small
\setlength{\tabcolsep}{8pt} 
\renewcommand{\arraystretch}{1.1}

\begin{tabular}{lcccc}
\toprule
\textbf{Condition} & \textbf{Easy}$^{\dagger}$ & \textbf{Middle} & \textbf{Hard}$^{\ddagger}$ & \textbf{Regime} \\
\midrule
DRMI-Landscape    & 63.5\% & 20.8\% & 15.7\% & Polarized \\
DRMI-Animal       & 49.4\% & 17.0\% & 33.6\% & Diffuse   \\
\arrayrulecolor{lightgrayline} 
\midrule
\arrayrulecolor{black}

Metaformer-Landsc.& 72.4\% & 20.9\% & ~~6.7\% & Plateau   \\
Metaformer-Animal & 74.9\% & 20.2\% & ~~4.9\% & Plateau   \\
\bottomrule
\multicolumn{5}{l}{\footnotesize $^{\dagger}$Agreement $\geq 0.8$. \;$^{\ddagger}$Agreement $\leq 0.6$.}
\end{tabular}
\end{table}

\section{Discussion}
\label{sec:discussion}
This study found that human-aligned ranking is both necessary and achievable under strict budget constraints. In this section, we analyze the structural causes of metric--human misalignment, explain why feature-based active learning struggles in this domain, and characterize the difficulty regimes where MetaRanker offers the greatest advantage. Finally, we discuss implications for perceptually grounded metalens co-design.

\noindent\textbf{Metric--Human Misalignment in Metalens Imaging.}
Across our experiments, standard IQA metrics exhibit weak or even negative correlations with human-derived rankings in the metalens domain (Section~\ref{sec:results_metrics}). While content diversity can amplify this discrepancy, it does not explain it: in the controlled Sim setting, where scene content is fixed and only optical degradation varies, full-reference metrics (PSNR, SSIM, LPIPS) remain poorly aligned with human rankings (Kendall's $\tau < 0.15$), and only NIQE achieves moderate correlation.

We argue that the root cause is structural. A t-SNE analysis of ResNet50 embeddings reveals that metalens-specific aberrations (e.g., chromatic misregistration, spatially varying blur,
and contrast loss) are entangled with conventional distortions such as noise or compression. Under such entanglement, proximity in feature space does not reliably correspond to proximity in perceived semantic interpretability. NSS features separate metalens outputs from the natural-image manifold more clearly, but they primarily capture statistical naturalness rather than recognizability. This misalignment has implications beyond evaluation. Any model-selection or optimization pipeline that uses these metrics as proxies for human interpretability inherits the same objective mismatch, potentially converging to numerically favorable solutions that do not preserve the semantic structures that matter for deployment utility. This concern is directly reflected in our simulation analysis.

\textbf{Implications for Feature-Based Active Preference Learning.}
The representational mismatch above directly affects feature-based active learning strategies. GURO identifies informative pairs using contextual embeddings, implicitly assuming that feature proximity reflects proximity in the target quality ordering. When this assumption fails---as the ResNet50 embedding analysis confirms---the acquisition policy can repeatedly select pairs that are close in embedding space yet uninformative for ordering by semantic interpretability. This mechanism explains the pronounced oscillations and early plateaus observed in simulation (Fig.~\ref{fig:simulation_annotation}) and the instability of feature-based methods across random seeds.

MetaRanker is designed to reduce dependence on external feature geometry. Its acquisition function (Eq.~\ref{eq:acquisition}) operates on uncertainty internal to the probabilistic rating model, rather than on distances in a pretrained embedding space. The VLM-based semantic prior plays a complementary role: it constrains candidate generation by grouping images with similar semantic-consistency scores, increasing the likelihood of near-boundary comparisons early. Notably, the benefit arises less from substituting a stronger backbone per se, and more from using a task-relevant semantic-consistency signal to narrow the search space in a way that does not require a well-calibrated embedding geometry for metalens artifacts.




\textbf{Regime-Dependent Benefits of Active Ranking.}
In this small expert-panel setting, Kendall's $\tau$ is the most appropriate primary IRR criterion for our main analysis. Each condition is evaluated by only seven annotators, with one session per annotator, so we avoid relying on strong assumptions about calibrated score magnitudes or stable rank-distance scales across raters. Kendall's $\tau$ measures pairwise concordance between induced rankings, matching both the pairwise annotation primitive and the active-ranking objective. Spearman's $\rho$ is retained as a complementary rank-consistency measure because it summarizes global rank displacement, but we treat it as secondary to the pairwise-order agreement captured by Kendall's $\tau$. We therefore focus our main analysis on these two rank-based IRR metrics: Kendall's $\tau$ for ordinal pairwise agreement and Spearman's $\rho$ for global rank consistency.

Although MetaRanker achieves the highest Kendall's $\tau$ across all conditions, the magnitude of improvement depends strongly on the baseline and condition. Relative to Bayesian CJ, the gain ranges from a modest margin on DRMI Animal to a large gap on DRMI Landscape (Section~\ref{sec:metric_ali}). We attribute this variation to how the adaptive volatility mechanism (Eq.~\ref{eq:self_correct}) interacts with the condition-specific difficulty structure (Table~\ref{tab:difficulty_advantage}).

In the \emph{polarized} regime of DRMI Landscape, hard pairs concentrate near a narrow decision boundary. When human feedback contradicts the model's current belief, the volatility mechanism inflates uncertainty for the relevant items, and the acquisition function revisits them. Because the informative region is compact, the feedback loop repeatedly allocates queries where information gain is highest. This yields a large Kendall gap against Bayesian CJ (Ours $0.786$ vs.\ $0.450$), while the stronger LBPS-EIC baseline narrows the margin in this condition.

In the \emph{plateau} regimes of Metaformer, most pairs are unambiguous (Easy $>72\%$, Hard $<7\%$), so large prediction errors and thus volatility reactivation occur infrequently. With limited near-boundary mass, the marginal benefit over simpler uncertainty-aware baselines is naturally smaller (Metaformer Animal: Ours $0.794$ vs.\ Bayesian CJ $0.612$). The \emph{diffuse} regime of DRMI Animal lies between these extremes: informative pairs exist but are broadly distributed, requiring wider exploration and yielding a moderate advantage. Overall, MetaRanker provides the largest gains when the quality landscape contains a meaningful yet localized boundary region, which commonly arises in metalens design spaces near recognizability thresholds.

\textbf{Objective Alignment and Implications for Co-Design.}
Our simulation experiments reveal a complementary insight. Under a consistent PSNR-based simulation target, MetaRanker and Bayesian CJ converge to similar Kendall's $\tau$.
When the target signal is internally consistent, asymptotic performance is determined more by the validity of the target than by small differences in acquisition design; crucially, this distortion-based proxy is structurally misaligned with human perception in the metalens domain (Table~\ref{tab:metric_vs_human_combined}).
High sample efficiency under such an objective mismatch leads to fast convergence toward the \emph{wrong target}. This motivates caution for end-to-end metalens pipelines that optimize pixel-fidelity proxies (e.g., PSNR/SSIM) as primary objectives~\cite{sitzmann2018end,tseng2021neural}, since they may reduce pixel-wise distortion without improving semantic interpretability.
Importantly, this paper does not present closed-loop co-design. Rather, we position MetaRanker as addressing the evaluation bottleneck required for perceptually aligned objectives. This framing follows established precedents: LPIPS~\cite{zhang2018unreasonable} was first validated as an evaluation metric before being adopted as a loss, and preference-based reward modeling~\cite{christiano2017deep} preceded its integration into RLHF pipelines~\cite{ouyang2022training}. Integrating MetaRanker into a co-design loop is technically feasible, e.g., by training a differentiable surrogate loss from rankings or by using the ranking as a fitness signal in derivative-free optimization~\cite{zini2025bayesian}, but empirical validation of such closed-loop integration remains future work.

\textbf{Limitations.}
This study has constraints that limit the generality of its conclusions. We evaluated two metalens settings (DRMI and Metaformer); because optical severity is confounded with dataset identity, relationships among severity, difficulty regime, and ranking advantage should be interpreted as suggestive rather than causal. The limited annotation data ($n{=}7$ sessions per dataset--category--method condition) restricts statistical resolution and increases sensitivity to outlier judgments, motivating our emphasis on descriptive bootstrap intervals and effect sizes rather than fine-grained $p$-values. Methodologically, gaps remain between simulation proxies and real-world deployment: ablations rely on synthetic rankings, and the simulation is driven by the PSNR metric, which is misaligned with human interpretability (Table~\ref{tab:metric_vs_human_combined}). Finally, MetaRanker is validated as an evaluation protocol; demonstrating end-to-end closed-loop integration remains out of scope.

\textbf{Future Work.}
Future work will address these limitations along three axes. First, we will extend evaluation to additional metalens systems spanning intermediate severity levels and characterize budget sensitivity via sweeps over $B \in [N, 5N]$, including studies with non-expert annotators. Second, we will close the loop between evaluation and optimization: as discussed in Section~\ref{sec:discussion}, this could involve training a differentiable surrogate model using pairwise preferences from MetaRanker (analogous to reward modeling in RLHF~\cite{christiano2017deep}) or using the ranking directly as a fitness signal in gradient-free optimization~\cite{zini2025bayesian}. In parallel, we will scale the VLM prior to larger candidate pools ($N{>}100$) and benchmark stronger foundation models to better characterize the trade-off between prior quality and annotation reduction. Third, we will explore transferring the same uncertainty-aware ranking protocol to related diffractive/holographic imaging settings where field-dependent aberrations and metric misalignment are also common.

\section{Limitations and Ethical Considerations}
\label{sec:limitations_ethics}
The DRMI dataset is an image dataset captured using a metalens that we independently designed and fabricated (Figshare, open access~\cite{seo2024deep}); the MetaFormer dataset is a simulated dataset~\cite{lee2024aberration}. Both are publicly available. No personally identifiable or sensitive data were collected from annotators; the study involved only image-quality preference judgments under the protocol in Section~\ref{sec:protocol_stat}.

\section{Conclusion}
\label{sec:conclusion}

MetaRanker addresses the prohibitive $O(N^2)$ annotation bottleneck in metalens image quality evaluation by introducing a "supervision-format shift" that formulates semantic interpretability as a human-grounded ranking problem rather than a distortion-metric optimization target. MetaRanker uses a fixed budget of only $B=3N$ pairwise comparisons per session, corresponding to an approximately 80\% reduction relative to exhaustive pairwise evaluation for $N=30$. Under this constrained labeling budget, MetaRanker achieves the top Final Rank across all animal and landscape conditions in both the DRMI and MetaFormer datasets (Table~\ref{tab:human_benchmark_stacked}). These results confirm that reliable human-aligned preference supervision can be efficiently obtained under strict annotation constraints.

A central design principle is that the VLM prior is used only as an imperfect sampling guide, not as a supervisory label or rating initializer. This flat-start strategy prevents anchoring bias, keeping human comparisons as the primary supervision signal throughout the ranking process. When observed human feedback contradicts the model's current preference prediction, the adaptive volatility mechanism raises uncertainty for the relevant items and redirects subsequent acquisition toward unresolved comparisons. This self-correcting behavior allows MetaRanker to benefit from "weak" AI semantic guidance without anchoring the final ranking to potentially hallucinated VLM prior predictions.

Overall, our findings suggest that standard IQA metrics do not accurately reflect human perception in the metalens domain, especially when semantic recognizability is degraded by chromatic and field-dependent aberrations that cause "recognizability collapse." MetaRanker therefore provides a practical evaluation protocol for perceptually grounded metalens assessment. Integrating the resulting human-aligned rankings into reconstruction training or closed-loop optical co-design is a natural next step. While our preliminary reconstruction pilots demonstrate that supervising pipelines with MetaRanker rankings yields significant gains in downstream perceptual (LPIPS) and naturalness (NIQE) metrics, full end-to-end empirical validation of such optimization pipelines remains future work.

\begin{acks}
This work was supported by the National Research Foundation of Korea (NRF) grants funded by the Ministry of Science and ICT (MSIT) (RS-\allowbreak2024-\allowbreak00338048, RS-\allowbreak2024-\allowbreak00455720, RS-\allowbreak2026-\allowbreak25487796); the National Institute of Health (NIH) research project (2026-\allowbreak ER0904-\allowbreak00); the Advanced GPU Utilization Support Program and the High Performance Computing Support project (RQT-\allowbreak25-\allowbreak070083), both funded by MSIT; Hankuk University of Foreign Studies Research Fund of 2026; the Culture, Sports and Tourism R\&D Program through KOCCA funded by MCST (RS-\allowbreak2024-\allowbreak00332210); the Artificial Intelligence Graduate School Program, Hanyang University (RS-\allowbreak2020-\allowbreak II201373); and the Artificial Intelligence Semiconductor Support Program (IITP-\allowbreak2025-\allowbreak RS-\allowbreak2023-\allowbreak00253914), supervised by IITP and funded by the Korean government.
\end{acks}

\section*{GenAI Disclosure}
Generative AI tools were not used for algorithm development, data collection, human annotation, statistical analysis, or result generation. Claude was used for code debugging assistance, and GPT-5 was used for grammatical and stylistic refinement. The authors reviewed and verified all AI-assisted content and take full responsibility for the work.

\balance
\bibliographystyle{ACM-Reference-Format}
\bibliography{ref}


\appendix






\appendix
\section*{Appendix}
\addcontentsline{toc}{section}{Appendix}

\section{Ablation Study Details}
\label{app:ablation}

\subsection{Hyperparameter Configuration}
\label{app:hyperparams}

Table~\ref{tab:hyperparams} lists the exact values used in all
main experiments. Sensitivity to $\alpha$ and $\mathrm{RD}_0$
is higher than for other parameters because these govern the
adaptive volatility correction, which is designed to buffer
against human label noise; $\kappa$ and $\theta$ are stable
across a wide range.

\begin{table}[h]
\centering
\small
\setlength{\tabcolsep}{4pt}
\renewcommand{\arraystretch}{1.1}
\caption{\textbf{Hyperparameter values used in all experiments.}
$\Delta\tau$ reports the maximum deviation from the default
across the one-factor sweep range shown.}
\label{tab:hyperparams}
\begin{tabular}{l l l c}
\toprule
\textbf{Symbol} & \textbf{Default} & \textbf{Sweep range} & $\boldsymbol{\Delta\tau}$ \\
\midrule
$\alpha$              & 0.5   & \{0.1, 0.3, 0.5, 1.0\}       & 0.130 \\
$\theta$              & 0.05  & \{0.01, 0.05, 0.10, 0.20\}   & 0.070 \\
$\mathrm{RD}_0$       & 350   & \{150, 250, 350, 500\}        & 0.118 \\
$\sigma_0$            & 0.06  & \{0.02, 0.06, 0.10, 0.20\}   & 0.019 \\
$\lambda$             & 0.8   & \{0.2, 0.5, 0.8, 1.5\}       & 0.038 \\
$\kappa$              & 100   & \{50, 100, 200, 400\}         & 0.025 \\
$\tau_{\text{auto}}$  & 0.85  & \{0.80, 0.85, 0.90, 0.95\}   & 0.014 \\
\bottomrule
\end{tabular}
\end{table}

The ablation study results are summarized in Table~\ref{tab:ablation_summary}. The simulation used a synthetic Bradley--Terry ground truth with $N=600$ items.

\textit{H1: Role of VLM Prior (Initialization vs.\ Sampling Guide).}
The results demonstrate that the VLM prior is most effective when used as a \emph{sampling guide} rather than for \emph{initialization}.
The configuration \textbf{Sampling guide (Flat start, $s=0$)} achieves the highest performance ($\tau=0.804$).
In contrast, directly initializing ratings with VLM scores (\textbf{Warm-start only, $s=100$}) yields a lower $\tau$ ($0.785$) than even the baseline ($0.797$).
This indicates that biasing the initial state with imperfect VLM predictions creates a ``bad prior'' that the ranking algorithm struggles to unlearn under a limited budget.
Instead, the sampling guide strategy uses the VLM solely to identify candidate pairs with similar scalar prior scores, e.g., small $|s_i-s_j|$ or adjacent ranks under $s_i$. By prioritizing such pairs for comparison, the algorithm filters out uninformative ``obvious''
pairs (e.g., severe blur vs.\ sharp) and focuses the budget on likely decision boundaries from the start, all while maintaining an unbiased flat initialization ($r=1500$) for the rating engine itself.

\textit{H2 \& H3: Necessity of Uncertainty-Awareness.}
Hypothesis H2 confirms that active acquisition is critical; \textbf{Random} selection performs significantly worse ($\tau=0.661$) compared to \textbf{Uncertainty} or \textbf{Hybrid} strategies ($\tau\approx0.730$), validating the efficiency of targeting decision boundaries.
Furthermore, H3 highlights the importance of modeling rating uncertainty explicitly.
\textbf{Elo}, which lacks a dynamic uncertainty parameter ($RD$), fails catastrophically ($\tau=0.298$), whereas \textbf{Glicko-2} maintains high performance.
This proves that the acquisition function relies heavily on the variance of the posterior distribution ($RD$) to identify informative pairs.

\textit{H4 \& H5: Optional Automation and Robustness.}
H4 shows that auto-comparison is not the primary source of the main performance gains. Conservative settings, such as Threshold 0.90 and Adaptive, remain comparable to the Off setting, whereas the more aggressive Threshold 0.85 setting reduces performance. Because this table reports final ranking accuracy rather than the number of saved human labels, we treat auto-comparison as an optional efficiency component rather than as a core supervision mechanism. H5 evaluates robustness to synthetic comparison-label noise. The results remain within a narrow range across tested noise levels, but they do not show a monotonic degradation pattern. Therefore, the slightly higher value at Noise $p=0.6$ should not be interpreted as evidence that noise improves ranking. Rather, the result supports a bounded robustness claim: adaptive volatility can prevent severe collapse when observed comparison outcomes are inconsistent with the model's current predictions. Adversarial VLM-prior corruption is evaluated separately in Section~\ref{app:corruption}.

\begin{table}[b!]
\centering
\footnotesize
\renewcommand{\arraystretch}{1.1}
\setlength{\tabcolsep}{4pt}
\caption{Ablation summary (mean$\pm$std). $N=600$, $B=1{,}800$. Note that $s$ denotes the warm-start spread: $s=0$ indicates a flat initialization ($r=1500$), while $s=100$ indicates a VLM-biased
initialization.}
\label{tab:ablation_summary}
\begin{tabular}{llc}
\toprule
Hypothesis & Method & Kendall $\tau$ \\
\midrule
\textbf{H1: Prior} & \textbf{Sampling guide (Flat start, $s=0$)} & \textbf{0.804$\pm$0.004} \\
(Init vs Guide) & Baseline (Flat start, $s=0$) & 0.797$\pm$0.002 \\
 & RD init (Flat start, $s=0$) & 0.784$\pm$0.009 \\
 & Full prior (Flat start, $s=0$) & 0.781$\pm$0.005 \\
 \cmidrule(l){2-3}
 & Warm-start only ($s=100$) & 0.785$\pm$0.009 \\
 & Warm-start + Sampling ($s=100$) & 0.767$\pm$0.008 \\
 & Warm-start + RD ($s=100$) & 0.741$\pm$0.004 \\
 & Full prior + warm-start ($s=100$) & 0.724$\pm$0.002 \\
\midrule
\textbf{H2: Acquis.} & Hybrid & 0.730$\pm$0.012 \\
 & Uncertainty & 0.729$\pm$0.009 \\
 & Boundary & 0.724$\pm$0.002 \\
 & Random & 0.661$\pm$0.009 \\
\midrule
\textbf{H3: Rating} & Glicko-2 (Adaptive) & 0.725$\pm$0.001 \\
 & Glicko-2 ($\tau=0.5$) & 0.724$\pm$0.002 \\
 & Elo ($K=32$) & 0.298$\pm$0.002 \\
 & Elo ($K=16$) & 0.153$\pm$0.001 \\
\midrule
\textbf{H4: Auto} & Threshold 0.90 & 0.724$\pm$0.002 \\
 & Adaptive & 0.724$\pm$0.002 \\
 & Off & 0.721$\pm$0.003 \\
 & Threshold 0.85 & 0.710$\pm$0.004 \\
\midrule
\textbf{H5: Label noise} & Noise $p=0.6$ & 0.730$\pm$0.017 \\
 & Clean ($p=0$) & 0.724$\pm$0.002 \\
 & Noise $p=0.4$ & 0.722$\pm$0.018 \\
 & Noise $p=0.8$ & 0.720$\pm$0.005 \\
 & Noise $p=0.2$ & 0.714$\pm$0.011 \\
\bottomrule
\end{tabular}
\end{table}


\section{VLM Prompting and Post-processing Details}
\label{app:vlm_details}

We summarize the VLM decoding configuration and deterministic post-processing rules
used to derive the VLM prior in Section~\ref{sec:prior_init}. The structured prompt is shown in Fig.~\ref{fig:vlm_prompt}; the score is computed from fixed decoded fields rather than free-form caption quality.

\subsection{Scope and design choice}
The \texttt{caption} field is logged for auditability but is not a primary signal in our score. The prior in Eq.~\ref{eq:prior_score} uses mapped visibility labels, object-set consistency, and normalized object count with $N_{\max}=8$ and equal weights unless otherwise stated.

\begin{table}[bht]
\centering
\small
\setlength{\tabcolsep}{4pt}
\renewcommand{\arraystretch}{1.08}
\caption{\textbf{VLM prompting configuration (fixed across datasets).}
We use an off-the-shelf VLM (LLaVA-7b via an Ollama API) and decode $K$ stochastic samples per image for
semantic stability estimation.}
\label{tab:vlm_settings}
\begin{tabular}{l l}
\toprule
\textbf{Item} & \textbf{Setting} \\
\midrule
VLM model & LLaVA-7b (Ollama backend) \\
Image encoding & JPEG, max side 768 px, quality 85 \\
Quality samples ($K$) & $K{=}2$ (default), fixed per experiment \\
Quality temperature & $T_{\text{VLM}}{=}0.7$ \\
Category temperature & $0.1$ (when category prompt is used) \\
Seeds & $17 + 101k$ for $k \in \{0,\dots,K{-}1\}$ \\
Max objects & up to 8 nouns (lowercased) \\
Truncation & caption: 400 chars; readable\_text: 200 chars \\
Fallback on parse failure & visibility=unclear, objects=[], confidence=0 \\
\bottomrule
\end{tabular}
\end{table}

\section{Descriptive Uncertainty Analysis}
\label{app:full_stats}

We summarize performance differences between MetaRanker and competing baselines using the seven human annotation sessions from Table~\ref{tab:human_benchmark_stacked}. For each (dataset, category, method) cell, the seven annotator-specific rankings yield $\binom{7}{2}=21$ pairwise inter-session correlations for Kendall's $\tau$ and Spearman's $\rho$. Mean Diff. is the mean correlation of MetaRanker minus that of the baseline; positive values indicate higher IRR for MetaRanker. We report paired non-parametric bootstrap CIs with 20{,}000 resamples and Cliff's $\delta$ effect sizes. Because the 21 correlations are not fully independent---the same annotator appears in multiple pairs---we interpret the intervals as descriptive uncertainty summaries rather than formal hypothesis tests.

\section{Preliminary Reconstruction-Adaptation Pilot}
\label{app:downstream}

As a preliminary qualitative check, we fine-tuned three DRMI reconstruction variants using MetaRanker-, PSNR-, or SSIM-ranked supervision and evaluated them on held-out DRMI images. The pilot suggests that MetaRanker-ranked supervision can improve perceptual metrics such as LPIPS and NIQE, whereas SSIM-ranked supervision tends to favor distortion-oriented scores. Because this experiment is limited in scale, we treat it as suggestive evidence only; large-scale reconstruction training and closed-loop co-design validation remain future work (Section~\ref{sec:discussion}).

\section{Prior Corruption Stress Test}
\label{app:corruption}

A potential concern is whether MetaRanker's self-correcting property holds when the VLM prior is severely miscalibrated. To stress-test this behavior, we construct adversarially corrupted priors by replacing a fraction $p$ of the prior scores with inverted scores, where $p{=}0$ denotes the original VLM prior and $p{=}1.0$ denotes a fully inverted prior. We then measure the final Kendall's $\tau$ under a controlled synthetic-oracle setting
($N{=}600$, $B{=}1{,}800$, 3 runs per level). We use PSNR only as a deterministic oracle for this stress test, not as a proxy for human perceptual quality.

The ranking degrades gracefully under moderate corruption ($\Delta\tau{=}{-}0.037$ at $p{=}0.5$), and remains usable even under a fully adversarial prior ($\Delta\tau{=}{-}0.078$). This supports a \emph{bounded} robustness claim: the prior is used only to guide early candidate pair generation, not as a supervisory label or rating initialization. When human feedback contradicts the prior-guided sampling trajectory, the prediction error inflates volatility and rating uncertainty (Eq.~\ref{eq:self_correct}),
redirecting subsequent queries toward uncertain items. A fully inverted prior is an extreme stress case unlikely in practice, since even poorly calibrated VLMs typically retain some semantic signal on non-degenerate images.

\begin{table}[t]
\centering
\caption{Descriptive comparison between MetaRanker and baselines.
Mean Diff. is Ours minus baseline; \textbf{bold} highlights a 95\% bootstrap CI excluding 0.}
\label{tab:full_bootstrap}
\resizebox{\columnwidth}{!}{%
\renewcommand{\arraystretch}{1.05}
\begin{tabular}{ll l c c}
\toprule
\textbf{Condition} & \textbf{Method} & \textbf{M}
    & \textbf{Mean Diff.\ [95\% CI]}
    & \textbf{Cliff's $\delta$\ [95\% CI]} \\
\midrule

\multirow{6}{*}{\shortstack[l]{DRMI\\Land.}}
  & \multirow{2}{*}{GURO}
    & $\tau$ & \textbf{+.363 [.180, .546]} & +0.72 [+0.44, +0.94] \\
  & & $\rho$ & \textbf{+.416 [.072, .760]} & +0.86 [+0.65, +1.00] \\
\cmidrule{2-5}
  & \multirow{2}{*}{Bayes CJ}
    & $\tau$ & \textbf{+.336 [.174, .498]} & +0.72 [+0.45, +0.92] \\
  & & $\rho$ & \textbf{+.226 [.167, .285]} & +0.95 [+0.85, +1.00] \\
\cmidrule{2-5}
  & \multirow{2}{*}{LBPS-EIC}
    & $\tau$ & +.013 [$-$.068, .094]        & +0.17 [$-$0.19, +0.51] \\
  & & $\rho$ & \textbf{$-$.053 [$-$.090, $-$.016]} & $-$0.58 [$-$0.84, $-$0.27] \\
\midrule

\multirow{6}{*}{\shortstack[l]{DRMI\\Animal}}
  & \multirow{2}{*}{GURO}
    & $\tau$ & \textbf{+.269 [.074, .464]}  & +0.57 [+0.27, +0.83] \\
  & & $\rho$ & \textbf{+.472 [.200, .744]}  & +0.82 [+0.54, +1.00] \\
\cmidrule{2-5}
  & \multirow{2}{*}{Bayes CJ}
    & $\tau$ & +.077 [$-$.058, .212]        & +0.27 [$-$0.09, +0.60] \\
  & & $\rho$ & $-$.016 [$-$.054, .022]      & $-$0.36 [$-$0.68, $-$0.00] \\
\cmidrule{2-5}
  & \multirow{2}{*}{LBPS-EIC}
    & $\tau$ & +.058 [$-$.073, .189]        & +0.06 [$-$0.30, +0.41] \\
  & & $\rho$ & +.016 [$-$.024, .056]        & $-$0.00 [$-$0.37, +0.35] \\
\midrule

\multirow{6}{*}{\shortstack[l]{MetaF\\Animal}}
  & \multirow{2}{*}{GURO}
    & $\tau$ & \textbf{+.522 [.290, .754]}  & +0.68 [+0.37, +0.93] \\
  & & $\rho$ & \textbf{+.799 [.392, 1.206]} & +0.72 [+0.43, +0.97] \\
\cmidrule{2-5}
  & \multirow{2}{*}{Bayes CJ}
    & $\tau$ & \textbf{+.182 [.058, .306]}  & +0.71 [+0.46, +0.90] \\
  & & $\rho$ & \textbf{+.035 [.003, .067]}  & +0.37 [+0.01, +0.70] \\
\cmidrule{2-5}
  & \multirow{2}{*}{LBPS-EIC}
    & $\tau$ & \textbf{+.286 [.137, .435]}  & +0.53 [+0.21, +0.81] \\
  & & $\rho$ & \textbf{+.184 [.131, .237]}  & +0.80 [+0.58, +0.96] \\
\midrule

\multirow{6}{*}{\shortstack[l]{MetaF\\Land.}}
  & \multirow{2}{*}{GURO}
    & $\tau$ & +.057 [$-$.025, .139]        & $-$0.01 [$-$0.37, +0.35] \\
  & & $\rho$ & \textbf{$-$.058 [$-$.090, $-$.026]} & $-$0.87 [$-$0.99, $-$0.69] \\
\cmidrule{2-5}
  & \multirow{2}{*}{Bayes CJ}
    & $\tau$ & \textbf{+.196 [.039, .353]}  & +0.59 [+0.27, +0.86] \\
  & & $\rho$ & +.023 [$-$.011, .057]        & +0.04 [$-$0.32, +0.40] \\
\cmidrule{2-5}
  & \multirow{2}{*}{LBPS-EIC}
    & $\tau$ & \textbf{+.183 [.087, .279]}  & +0.87 [+0.70, +0.98] \\
  & & $\rho$ & \textbf{+.064 [.021, .107]}  & +0.45 [+0.11, +0.76] \\
\bottomrule
\end{tabular}%
}
\end{table}

\begin{table}[h]
\centering
\small
\setlength{\tabcolsep}{8pt}
\renewcommand{\arraystretch}{1.1}
\caption{\textbf{Robustness to adversarial prior corruption.}
Final Kendall's $\tau$ degrades gracefully under moderate corruption, supporting
a bounded robustness claim for the self-correcting mechanism.}
\label{tab:corruption}
\begin{tabular}{cccc}
\toprule
\textbf{Corruption} $p$ 
    & \textbf{Initial} $\tau$ \textbf{(prior)} 
    & \textbf{Final} $\tau$ 
    & $\boldsymbol{\Delta}$ \textbf{from clean} \\
\midrule
0.0 (clean)      & $\phantom{-}0.170$ & $0.747$ & --- \\
0.5              & $\phantom{-}0.003$ & $0.710$ & $-0.037$ \\
1.0 (inverted)   & $-0.170$           & $0.669$ & $-0.078$ \\
\bottomrule
\end{tabular}
\end{table}

\end{document}